%% file: ms.tex
\documentclass[letterpaper, 10 pt, conference]{ieeeconf}  

\IEEEoverridecommandlockouts                              

\usepackage{graphics} 
\usepackage{epsfig} 
\usepackage{mathptmx} 
\usepackage{times} 
\usepackage{amsmath} 
\usepackage{amssymb}  
\usepackage{booktabs}
\usepackage{subcaption}
\DeclareMathOperator*{\argmax}{arg\,max}

\DeclareMathAlphabet{\mathcal}{OMS}{cmsy}{m}{n}

\title{\LARGE \bf
Learning multimodal representations for sample-efficient recognition of human actions*
}

\author{Miguel Vasco$^{1}$ \and Francisco S. Melo$^{1}$ \and David Martins de Matos $^{1}$ \and Ana Paiva$^{1}$ \and Tetsunari Inamura$^{2}$%
\thanks{*The first author acknowledges the FCT grant SFRH/BD/139362/2018.}%
\thanks{$^{1}$M.~Vasco, F.S.~Melo, D.M~de Matos and A.~Paiva are with INESC-ID and Instituto Superior T\'{e}cnico, University of Lisbon, Portugal. E-mail:
        {\tt\small miguel.vasco@gaips.inesc-id.pt} and {\tt\small\{fmelo,david.matos,ana.paiva\}@inesc-id.pt}.}
\thanks{$^{2}$T.~Inamura is with the National Institute of Informatics and the Department of Informatics, SOKENDAI (The Graduate University for Advanced Studies), 2-1-2 Hitotsubashi, Chiyoda-ku, Tokyo, Japan. Email: {\tt\small inamura@nii.ac.jp}}}

\begin{document}

\maketitle
\thispagestyle{empty}
\pagestyle{empty}

\begin{abstract}

Humans interact in rich and diverse ways with the environment. However, the representation of such behavior by artificial agents is often limited. In this work we present \textit{motion concepts}, a novel multimodal representation of human actions in a household environment. A motion concept encompasses a probabilistic description of the kinematics of the action along with its contextual background, namely the location and the objects held during the performance. Furthermore, we present Online Motion Concept Learning (OMCL), a new algorithm which learns novel motion concepts from action demonstrations and recognizes previously learned motion concepts. The algorithm is evaluated on a virtual-reality household environment with the presence of a human avatar. OMCL outperforms standard motion recognition algorithms on an one-shot recognition task, attesting to its potential for sample-efficient recognition of human actions. 

\end{abstract}

\input{body}



\bibliographystyle{IEEEtran}

\end{document}

%% file: body.tex
\setlength{\belowcaptionskip}{-5pt}

\section{Introduction}

Humans are able to interact with their environment in rich and diverse ways. Such richness and variety make it impossible to program an artificial agent that is able to recognize all possible actions performed by a human user. One common approach is to program agents to learn to recognize new human actions from demonstrations. However, it is unrealistic to assume that such learning will depend on large amounts of data, as required by many current learning algorithms. Instead, the agent should be able to learn and recognize novel actions from just a few demonstrations provided by the human.

To attain such efficient learning, the learning process should take into account the multimodal information provided by the human to create a rich representation of the novel action. However, the conventional methodology of learning human action representations considering only motion pattern data neglects the rich contextual background of the demonstration. This negligence results in a limited representation of the human action, hindering its recognition and introducing difficulties in the distinction between actions with similar motion patterns.

In this work, we address the problem of learning and recognizing human actions, from few demonstrations provided by a human in a household environment. We propose a novel representation for multiple demonstrations of a given action, named \textit{motion concept}. The motion concept encompasses a probabilistic motion primitive description of the motion patterns observed, augmenting it with their contextual background information, namely the location of the action and the objects used during the demonstrations. Moreover, the motion concept takes into account information provided directly through interaction with a human and allows the agent to reason on the importance of each contextual feature for its recognition.

Furthermore, we present the Online Motion Concept Learning (OMCL) algorithm, responsible for the creation of new motion concepts through interaction with a human user. The algorithm is able to recognize motion concepts from a single training demonstration and continuously update motion concepts as more demonstrations are provided. We evaluate the algorithm's performance on an offline "one-shot" motion recognition task, showing the importance of contextual information for the recognition of motion concepts built from a single training demonstration. The obtained results attest to the potential of OMCL for sample-efficient recognition of human actions.

\section{Related Work}

The question of learning motion and action representations has been addressed in literature, in part due to the widespread availability of low-cost motion sensing devices~\cite{lun2015survey}. Several representations have been proposed to model human action based on motion data. Xia \textit{et al.}~\cite{xia2012view} propose a view-invariant action representation based on histograms of 3D joint position, in relation to a fixed coordinate system, obtained from Kinect depth maps. The temporal evolution of these representations are modelled according to discrete HMMs. The authors in~\cite{yang2012eigenjoints} propose a novel feature for human action recognition based on the differences in the position of joints in the skeletal model of the human, employing a naive-bayes-nearest-neighbor (NBNN) classifier for recognition of the action classes. The authors in~\cite{ofli2014sequence} propose an interpretable representation of an action based on the sequence of joints in the skeletal model of the human which, at each time instant, are considered to be the most informative of a given demonstration. The informative criteria are based on predefined measures, such as the mean and variance of joint angle trajectories. Vemulapalli \textit{et al.}~\cite{vemulapalli2014human} model an action as a curve in the Lie group manifold. The curve of each action is generated based on a novel skeletal-based representation that explicitly models the geometric relationships between various body parts using rotations and translations in 3D space. However, all the presented representations of human actions are built solely resorting to motion data. As such, they neglect the rich contextual background of an action, which is fundamental for the distinction of action classes with similar motion patterns.

Moreover, several deep-learning frameworks have been recently proposed for motion and action recognition. Du \textit{et al.}~\cite{du2015hierarchical} proposed a hierarchical recurrent neural-network (RNN) framework for skeleton based action recognition, in which the human skeleton is divided accordingly to the human's physical structure. Multiple bidirectional RNNs (BRRN) are trained for each segmented section of the skeleton model, and their output is fused hierarchically by the upper-layers of the framework. Simonyan \textit{et al.}~\cite{simonyan2014two} propose a convolutional network architecture for action recognition in video, that incorporates spatial and temporal networks. While these architectures obtain impressive recognition results, their requirement of large amounts of training data make it unsuitable for the recognition of novel actions from few demonstrations.

Multimodal approaches to the creation of action representations have also been explored in literature. The authors in~\cite{wang2012mining} represent actions as an ensemble model and have proposed novel features for depth data which capture human motion and human-object interaction data from a demonstration. Using image data, Yao \textit{et al.}~\cite{yao2011human} learn action representations composed of attributes, words that describe the properties of human actions, and action-parts, the objects and poselets that are related to the actions. The authors in~\cite{ikizler2010object} model an action by integrating multiple feature channels from several entities (such as objects, scenes and people), extracted from video sequences. The representations are obtained through a "multiple instance learning" (MIL) model, where a given action label is associated with a group of instances.

\begin{figure}[h]
  \centering
    \includegraphics[width=0.49\textwidth]{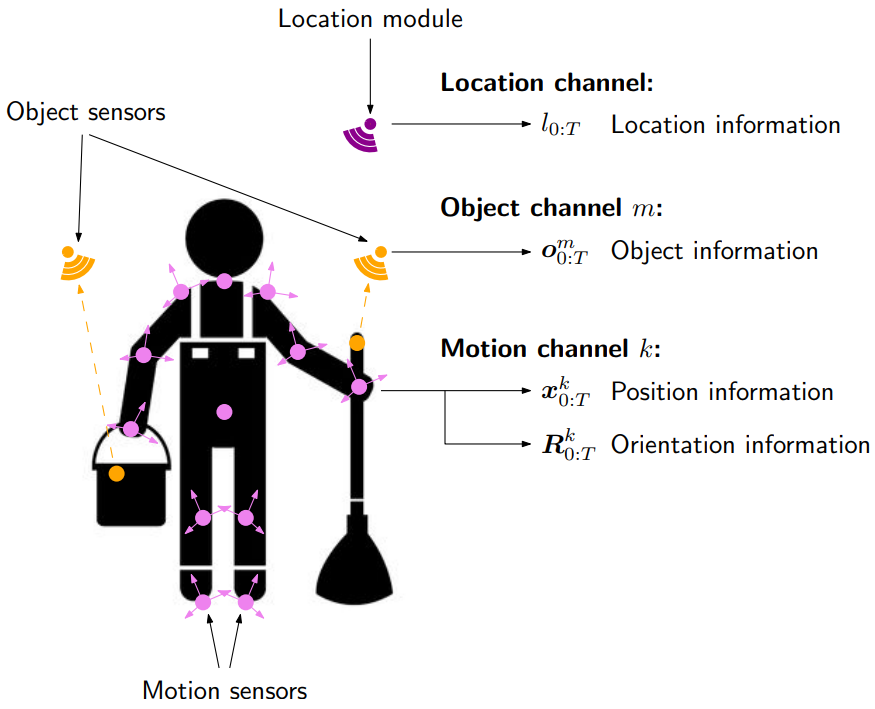}
    \caption{Depiction of the setup used throughout the paper. Sensors correspond
to input channels and provide streams of observations of length $T+1$}
    \label{I:setup}
\end{figure}

The importance of the action representation and recognition fields is attested by the extensive literature on the matter. Yet, the question regarding the creation of multimodal action representations from few demonstrations still requires addressing.

\section{Conceptual Representation of an Action}
\subsection{The setup}

We consider the following setup for learning from demonstration. A human user demonstrates an action, which may involve interaction with objects in the environment. The environment comprises a number of locations of interest, and the human user may be in any of these locations at the time of demonstration.

We assume that the environment is engineered with a number of sensors, providing information regarding the \textit{location} and \textit{pose} of the human user as well as the \textit{objects} that the user interacts with. In this work, we are not concerned with the actual sensing and admit that the system may include both "internal" sensors (e.g., data gloves to capture pose information, etc.) or external (e.g., cameras or optical trackers). In practical terms, the sensors act as input channels for the system, and as such we henceforth refer to sensors generally as a \textit{channels}. For example, a sensor deployed to provide object information is referred simply as an \textit{object channel}, and sensors deployed to track the human pose are referred as \textit{motion channels} (see Fig.~\ref{I:setup} for an illustration). Finally, the location of the user in the environment is provided by a dedicated sensing module, referred to as the \textit{location channel}. A demonstration by a human user yields a number of data streams arising
from the different input channels. In particular, \begin{itemize}

 \setlength\itemsep{1em}
    \item Each motion channel $k, k = 1, ..., K$ provides two streams of length $T$, $\boldsymbol{x}_{0:T}^k$ and $\boldsymbol{R}_{0:T}^k$, where each $\boldsymbol{x}_{t}^k$ indicates the position of a body element (joint, limb) at time step $t, t = 0, ..., T$, measured with respect to a common fixed world frame, and each $\boldsymbol{R}_{t}^k$ is a rotation matrix representing the orientation of that same body element at time step $t$;
    
    \item Each object channel $m, m = 1, ..., M$, provides one stream of length $T$, $\boldsymbol{o}_{0:T}^m$, where each individual observation $\boldsymbol{o}_{t}^m$ corresponds to a binary vector indicating the objects (from a predefined finite set of objects $\mathcal{O}$) that the user is interacting with at time step $t, t = 0, ..., T$, according to channel $m$;
    \item Finally, the location module provides a stream of length $T$, $l_{0:T}$, where $l_t$ indicates the location of the user at time step $t$. We assume that the location of the user takes values in a finite set $\mathcal{L}$ of possible locations. 
    
\end{itemize}

Learning a representation of an action will consist of taking the streams from the different input channels and compile them into a unique, compact representation that we refer to as a motion prototype, described in the continuation.

\subsection{Motion prototype}

The central constituent in our proposed action representation is the \textit{motion
prototype}, providing a compact representation for a single demonstration of an action by a human user. In particular, motion prototypes capture in a probabilistic manner motion information (extracted from the motion channels) and object and location information.

Formally, we represent a motion prototype as a tuple $P = (\boldsymbol{\tau}, \boldsymbol{\rho}, \lambda)$, where $(\boldsymbol{\rho}, \lambda)$ summarize the associated context information - namely object and location information - and $\boldsymbol{\tau}$ summarizes the motion observed in the demonstration. Specifically,

\begin{itemize}
 \setlength\itemsep{1em}

    \item $\boldsymbol{\rho} = \left\{ \rho_m, m = 1, ..., M \right\}$, where $M$ is the total number of object channels. For every object $o \in \mathcal{O}$,
    \begin{equation*}
    \rho_m(o) = \mathbb{P}\left[ \text{o}_{t,o}^m = 1, t = 0, ..., T\right] 
    \end{equation*}
    \item[] where o$_{t,o}^m$ is a random variable indication whether object $o$ was observed in object channel $m$ at time step $t$. In other words, in our proposed representation we assume that the observation of an object $o \in \mathcal{O}$ in channel $m$ at any moment during the human demonstration can be described probabilistically as a Bernoulli random variable with parameter $\rho^m(o)$.

    \item For every location $l \in \mathcal{L}$,
    \begin{equation*}
    \lambda(l) = \mathbb{P}\left[ \text{l}_t = l, t = 0, ..., T\right] 
    \end{equation*}
    \item[] where l$_t$ is a random variable indicating the location of the human demonstrator at time step $t$. In other words, in our representation we assume that the location of the human user during the demonstration can be described probabilistically as a categorical distribution with parameters $\lambda(l), l \in \mathcal{L}$.
    
\end{itemize}


Finally, we have that $\boldsymbol{\tau} = \left\{ \tau_k, k = 1, ..., K \right\}$, where $K$ is the number of motion channels and each $\tau_k$ is a sequence of \textit{motion primitives} $\left\{ \phi_n, n = 1, ..., N\right\}$. The concept of motion primitive has been widely explored both to describe animal motion and to represent robot motion~\cite{flash2005motor,kober2009learning}. For our purposes, a motion primitive $\phi_n$ is a probability distribution over the space of trajectories. In other words, given an arbitrary trajectory $\left(\boldsymbol{x}_{0:T}, \boldsymbol{R}_{0:T}\right)$,
\begin{equation*}
    \phi_n\left(\boldsymbol{x}_{0:T}, \boldsymbol{R}_{0:T}\right) = \mathbb{P}\left[ \boldsymbol{\mathrm{x}}_{0:T}^n = \boldsymbol{x}_{0:T}, \boldsymbol{\mathrm{R}}_{0:T}^n = \boldsymbol{R}_{0:T} \right] 
\end{equation*}

For the purpose of learning and recognition, it is convenient to treat an action not as comprising a single trajectory $(\boldsymbol{x}_{0:T}, \boldsymbol{R}_{0:T})$ but, instead, as a sequence of smaller trajectories,

\begin{equation*}
    \left\{ \left( \boldsymbol{x}_{0:t_1}, \boldsymbol{R}_{0:t_1} \right), \left( \boldsymbol{x}_{t_1:t_2}, \boldsymbol{R}_{t_1:t_2} \right), ..., \left( \boldsymbol{x}_{t_{N-1}: t_N}, \boldsymbol{R}_{t_{N-1}: t_N} \right)   \right\}
\end{equation*}

which are then encoded as a sequence of motion primitives $\left\{ \phi_n, n = 1, ..., N\right\}$, each $\phi_n$ providing a compact description of \newline($\boldsymbol{x}_{t_{n-1}: t_n}$, $\boldsymbol{R}_{t_{n-1}: t_n})$. Each motion primitive $\phi_n$ is selected among a library $\Phi$ of available motion primitives to maximize the likelihood of the observed trajectory, i.e.,
\begin{equation}
    \phi_n = \argmax_{\phi \in \Phi} \phi\left( \boldsymbol{x}_{t_{n-1}: t_n}, \boldsymbol{R}_{t_{n-1}: t_n} \right)
    \label{EQ:selection_primitive}
\end{equation}

\begin{figure}
  \centering
    \includegraphics[width=0.49\textwidth]{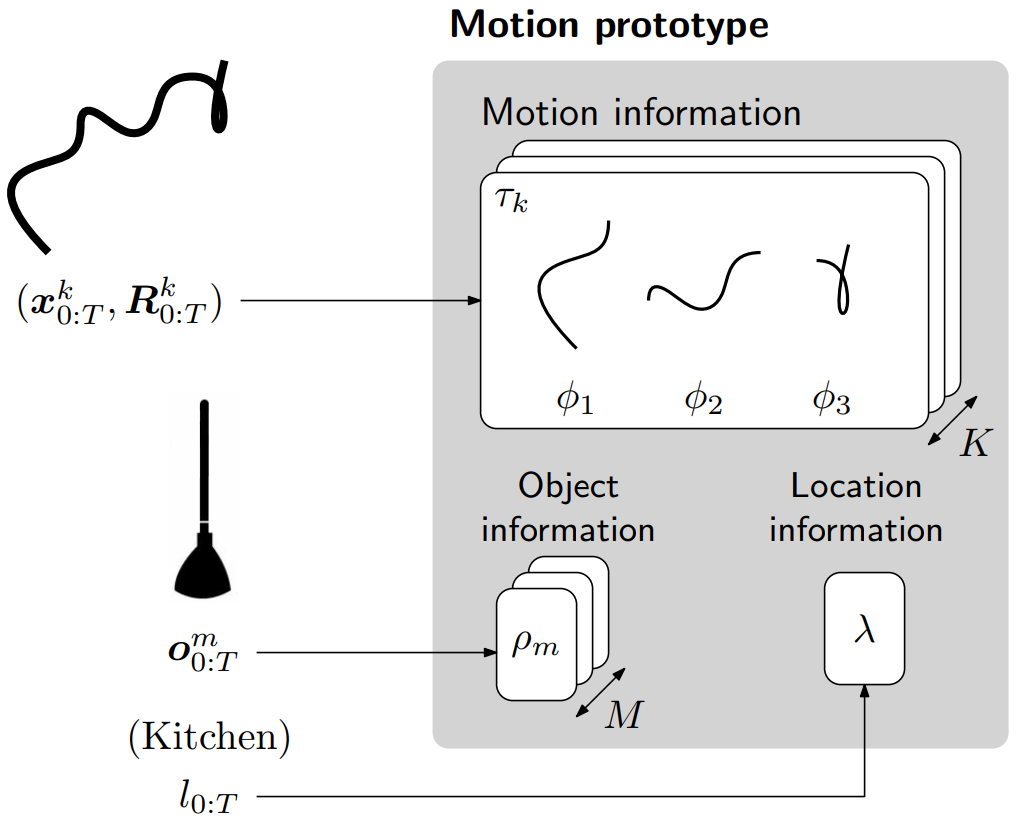}
    \caption{Summary representation of a motion prototype.}.
    \label{I:prototype}
\end{figure}

Summarizing, a motor prototype compactly encodes a demonstration of an action in the form of a tuple $(\boldsymbol{\tau}, \boldsymbol{\rho}, \lambda)$, where $\boldsymbol{\tau}$ is a collection of trajectories (one for each motion channel), each represented as a sequence of motor primitives; $ \boldsymbol{\rho}$ is a collection of probability distributions (one for each object channel), describing how the human interacted with each object in the environment; and $\lambda$ is a probability distribution describing where the human was located during the demonstration (see Fig.~\ref{I:prototype}).

\subsection{Motion concept}

It is possible for a single action to be performed in multiple different ways. A motion prototype, while providing a convenient representation for a single demonstration (and corresponding context), is insufficient to capture the diversity that a broader notion of "action" entails.


We introduce \textit{motion concept} as a higher-level representation of an action. A motion concept seeks to accommodate the different ways by which an action may be performed while, at the same time, encode distinctive aspects that are central in recognizing such action. For example, to distinguish actions such as "waving goodbye" and "washing a window", it is important to note that the latter involves interaction with an object (such a sponge) while the first does not.

\begin{figure}[h]
  \centering
    \includegraphics[width=0.49\textwidth]{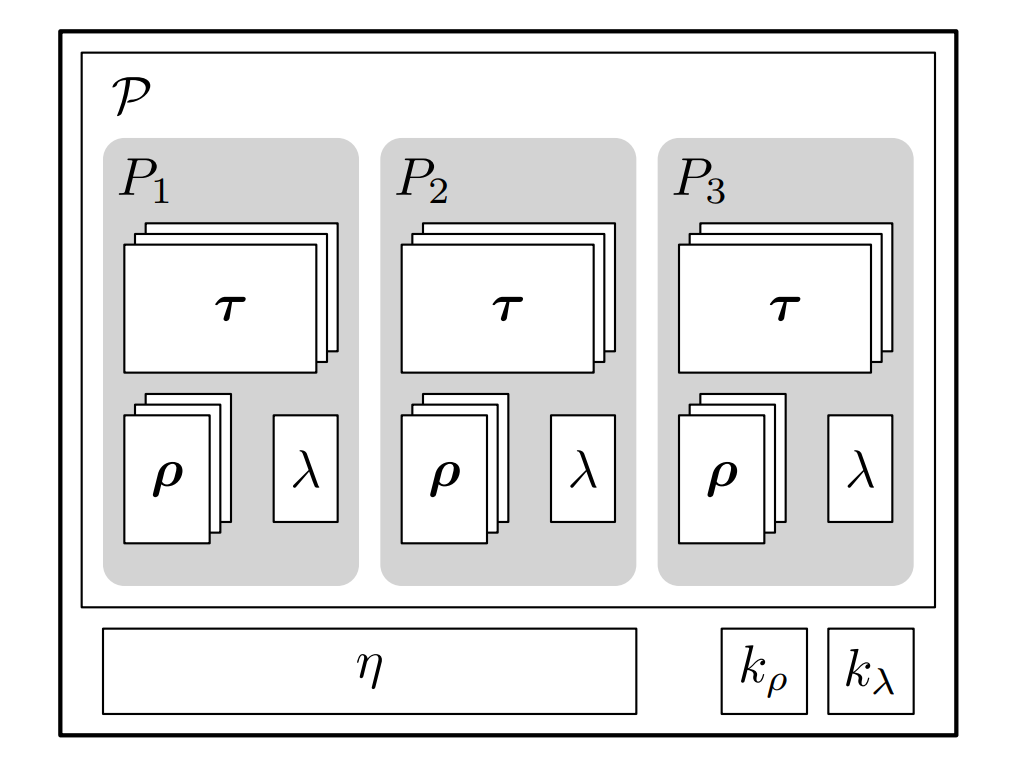}
    \caption{Schematic representation of a motion concept.}.
    \label{I:motion_concept}
\end{figure}

Formally, a motion concept consists of a tuple $ \mathcal{M} = \left( \mathcal{P}, \eta, k_{\rho}, k_{\lambda} \right)$ used to represent some action $a$, where
\begin{itemize}
 \setlength\itemsep{1em}
    \item $\mathcal{P} = \left\{ P_1, ..., P_\ell \right\}$, where each $P_i$ is a motion prototype describing one possible way by which the action $a$ can be performed;
    \item $\eta$ is a \textit{designation} (a "name") provided by the user to refer to the action $a$ - for example, it may consist of a label for $a$ or an utterance that corresponds to the spoken designation of $a$;
    \item $k_p$ and $k_{\lambda}$ are two constants used to weight the importance of object information and location information in recognizing the action $a$.
\end{itemize}

We schematically depict a motion concept in Fig.~\ref{I:motion_concept}.

\section{Learning motion concepts}

A motion concept provides a multimodal representation of a given action demonstrated by a human. However, in order for an agent to learn motion concepts from demonstration a novel algorithm is required. For that purpose, we now introduce the OMCL (Online Motion Concept Learning) algorithm, designed to construct motion concepts from the demonstration data provided by a human user. 

The OMCL algorithm proceeds can roughly be understood as working on two different abstraction levels. At a lower level, OMCL takes the data from a single demonstration and constructs a motion prototype from such data. At a higher level, OMCL combines information from multiple demonstrations to build a motion concept that potentially contains multiple motion prototypes. We now detail OMCL at each of these two abstraction levels.

\subsection{Learning a motion prototype from data}

Given a demonstration $\left(\boldsymbol{x}_{0:T}, \boldsymbol{R}_{0:T}\right)$, OMCL starts by segmenting the single trajectory into multiple sub-trajectories
\begin{equation}
    \left\{\left( \boldsymbol{x}_{0: t_{1}}, \boldsymbol{R}_{0: t_{1}} \right), ..., \left( \boldsymbol{x}_{t_{N-1}: t_N}, \boldsymbol{R}_{t_{N-1}: t_N} \right) \right\}
    \label{EQ:Segmentation}
\end{equation}

which can be achieved using any segmentation method from the literature - OMCL is agnostic to the particular segmentation method used. OMCL uses online kernel density estimation\footnote{In our implementation, we use the XOKDE++ algorithm from~\cite{ferreira2016fast}} to construct new motion primitives from sub-trajectory data. The list of segmented sub-trajectories (Eq.~\ref{EQ:Segmentation}) is used to update the previous set $\Phi$ of available motion primitives, after which each sub-trajectory $\left( \boldsymbol{x}_{t_{n-1}: t_n}, \boldsymbol{R}_{t_{n-1}: t_n} \right)$ is evaluated against the updated $\Phi$ and a primitive $\phi_n$ is selected accordingly to Eq.~\ref{EQ:selection_primitive}. The resulting sequence of motion primitives, $\left\{ \phi_1, ..., \phi_N \right\}$, corresponds to the trajectory representation $\tau$, as described in the previous section. Such procedure is repeated for all motion channels.

As for $\boldsymbol{\rho}$, we use standard maximum likelihood estimation to compute the parameters $\rho(o), o \in \mathcal{O}$, from the data $\boldsymbol{o}_{0:T}$, repeating this procedure for all object channels. Finally, we also use maximum likelihood estimation to compute the parameters $\lambda(l), l \in \mathcal{L}$, from the data $l_{0:T}$.

\subsection{Building a motion concept}
\label{S:Build_MC}

From a provided demonstration of action $a$, the OMCL algorithm learns a motion prototype  $P_a = (\boldsymbol{\tau}_a, \boldsymbol{\rho}_a, \lambda_a)$. If $P_a$ is the first motion prototype of action $a$ provided, a new motion concept $\mathcal{M}_a$ is built through the following procedure:
\begin{itemize}
 \setlength\itemsep{1em}
    \item The motion prototype is added to the empty list of prototypes $\mathcal{P}$,
    \begin{equation*}
        \mathcal{P} = \left\{ P_a\right\}
    \end{equation*}
    
    \item The importance weights $k_{\rho}, k_{\lambda}$ are initialized to predetermined values,
    \begin{equation*}
        k_{\rho} = k_{\rho,0},\quad k_{\lambda} = k_{\lambda,0}
    \end{equation*}
    
    \item The human is queried for the designation $\eta$ of the action.
    
\end{itemize}
The novel motion concept $\mathcal{M}_a$ is added to the current list of built motion concepts $\Sigma = \left\{\mathcal{M}_1, ..., \mathcal{M}_A \right\}$, where $A$ is the total number of action classes previously demonstrated. If the motion prototype $P_a$ concerns an action $a$ previously demonstrated, it is then used to update the respective motion concept $\mathcal{M}_a$, following:
\begin{itemize}
 \setlength\itemsep{1em}
 
    \item The motion prototype is added to the list of prototypes $\mathcal{P}$,
    \begin{equation*}
        \mathcal{P} = \left\{ P_1, ..., P_a \right\}
    \end{equation*}
    \item The contextual information of the motion prototype $(\boldsymbol{\rho}_a, \lambda_a)$ is used to update the values of the importance weights $k_{\rho}, k_{\lambda}$. If, for the majority of motion prototypes $P_i =  (\boldsymbol{\tau}_i, \boldsymbol{\rho}_i, \lambda_i) \in \mathcal{P}$:
     \begin{equation*}
        \argmax_{o \in \mathcal{O}} \rho_{i,m}(o) = \argmax_{o \in \mathcal{O}} \rho_{a,m}(o),  \forall m \in M
    \end{equation*}
    
    we increase the value of $k_{\rho}$ by a percentage $\alpha_k$ of its value. Otherwise, we decrease it by the same percentage. The same procedure is applied for the update of the $k_{\lambda}$ weight.

\end{itemize}

\section{Recognizing Motion Concepts}
\label{S:OMCL:Rec}

Beyond creating and updating motion concepts, the OMCL algorithm is also responsible for the recognition of previously observed actions and for the assessment of the novelty of previously unobserved action classes. Given a motion prototype from an unknown action $P_\ast = (\boldsymbol{\tau}_\ast,$ $\boldsymbol{\rho}_\ast,$ $\lambda_\ast)$, OMCL compares $P_\ast$ with the prototypes contained in every motion concept in the current set $\Sigma$. The cost of assigning $P_\ast$ to the motion concept $\mathcal{M}_i$ is given by:
\begin{equation}
    \mathcal{C}(\mathcal{M}_i, P_\ast) = \mathcal{C}_{\boldsymbol{\tau}}(\mathcal{P}, \boldsymbol{\tau_\ast}) + k_{\rho} \mathcal{C}_{\boldsymbol{\rho}}(\mathcal{P},\boldsymbol{\rho}_\ast) + k_{\lambda} \mathcal{C}_{\lambda}(\mathcal{P}, \lambda_\ast )
    \label{EQ:total_comp_cost}
\end{equation}
where,
\begin{itemize}
    
    \setlength\itemsep{1em}

    \item $\mathcal{C}_{\boldsymbol{\tau}}(\mathcal{P}, \boldsymbol{\tau_\ast})$ is the average cost of comparing the sequence of motion primitives in $\boldsymbol{\tau}_\ast$ to the sequence of motion primitives $\tau$ of every motion prototype $P \in \mathcal{P}$ of $\mathcal{M}_i$. To compute this cost, we use dynamic time warping~\cite{muller2007dynamic} between the sequences of each motion channel, with 0-1 loss;
    
    \item $\mathcal{C}_{\boldsymbol{\rho}}(\mathcal{P},\boldsymbol{\rho}_\ast)$ is the average distance between $\boldsymbol{\rho}_\ast$ and the collection of object probability distributions $\boldsymbol{\rho}$ of every motion prototype $P \in \mathcal{P}$ of $\mathcal{M}_i$. The algorithm is agnostic to the type of metric used to compute the distance between distributions;
    
    \item $ \mathcal{C}_{\lambda}(\mathcal{P}, \lambda_\ast)$ is the average distance between $\lambda_\ast$ and the location probability distributions $\lambda$ of every motion prototype $P \in \mathcal{P}$ of $\mathcal{M}_i$;
    
    \item $k_{\rho}, k_{\lambda}$ are the object and location information weights of $\mathcal{M}_i$.
     
\end{itemize}

For recognition purposes, the cost is computed for all motion concepts available in $\Sigma$ and $P_\ast$ is assigned to the motion concept $\mathcal{M}_R \in \Sigma$ with the lowest assignment cost $\mathcal{C}_{R}$.

However, the question of the possible class-novelty of the demonstration still requires addressing. Therefore, subsequently, OMCL determines if $P_\ast$ belongs to the assigned motion concept or if it belongs to a novel action class. The decision takes into account the average wrong cost $\mathcal{C}_{W}$ of assigning $P_\ast$ to the remaining motion concepts $\mathcal{M} \in \Sigma \setminus \mathcal{M}_R$. In the case,
\begin{equation}
    | \mathcal{C}_{R} - \mathcal{C}_{W}| \geq \Delta_{\mathcal{C}} \times \mathcal{C}_{R}
\label{Eq:novel_mc}
\end{equation}
the provisional assignment of $P_\ast$ to $\mathcal{M}_R$ is confirmed, where $\Delta_{\mathcal{C}}$ is a predefined constant. Otherwise, the assignment can be considered quasi-random, and $P_\ast$ will be used to build a new motion concept (as detailed in the previous section).

\section{Evaluation}

\subsection{Experimental Setup}

\begin{figure}
  \centering
    \includegraphics[width=0.45\textwidth]{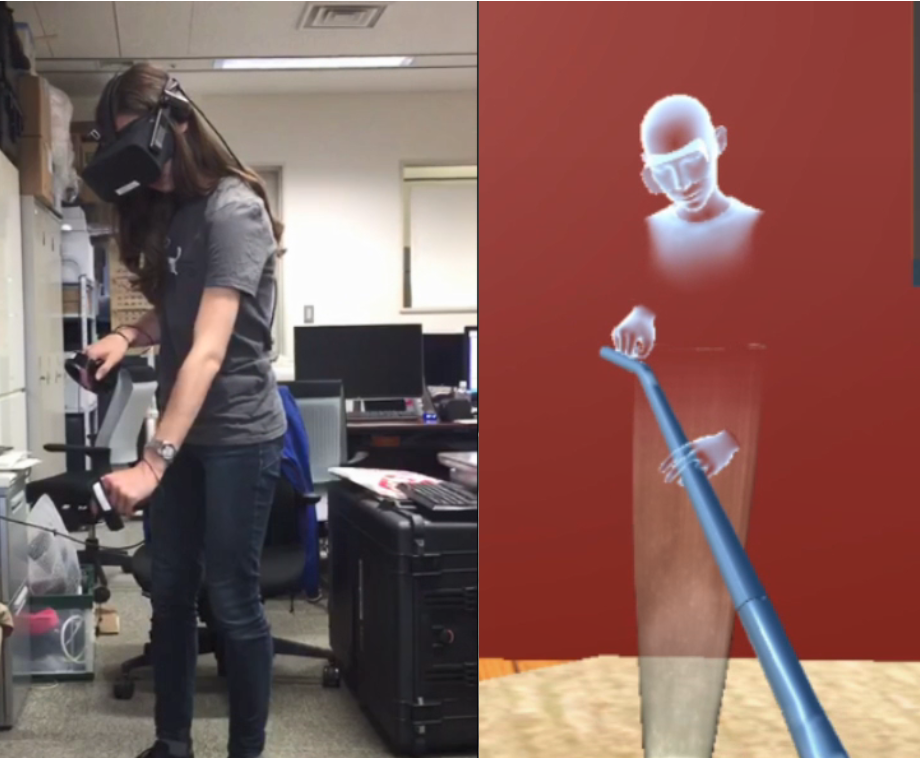}
    \caption{Participant demonstrating an example of the "Vacuum-clean" action, along with the corresponding VR avatar.}
    \label{I:vr_interaction}
\end{figure}

The evaluation of the OMCL algorithm was performed in a virtual-reality (VR) environment. The user interacts with the VR environment using a Oculus Rift headset and hand motion controllers, as shown in Fig.~\ref{I:vr_interaction}. Thus, in this setup, the number of motion channels $K$ is equal to the number of object channels $M$, with $K = M = 3$.

\begin{figure}
  \centering
    \includegraphics[width=0.4\textwidth]{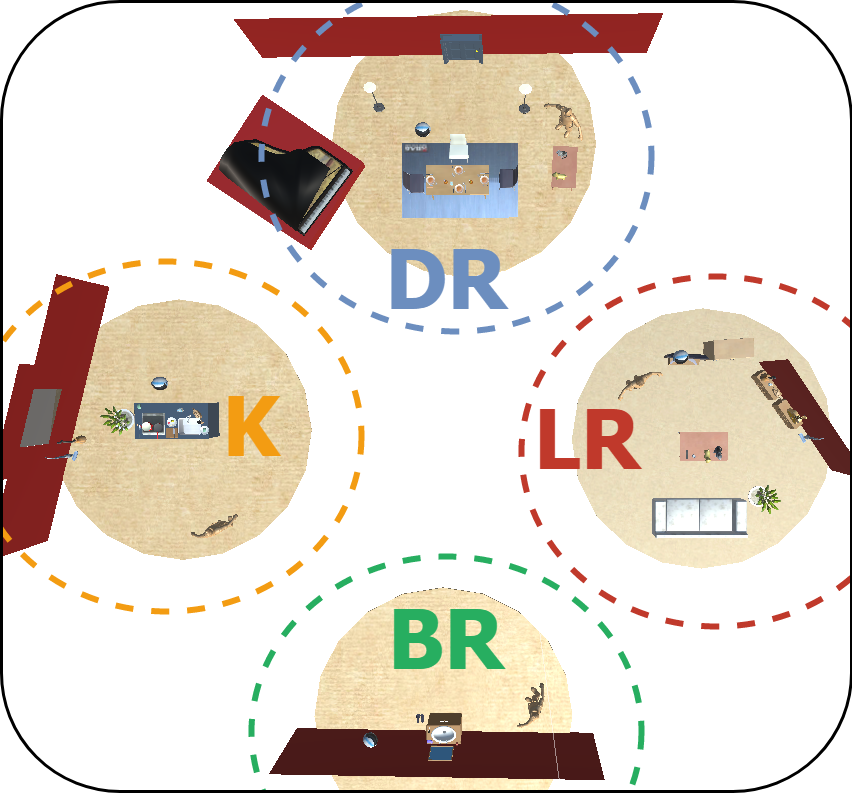}
    \caption{Virtual household environment with the area of every discrete location discriminated: "DR"-Dining Room, "K"-Kitchen, "BR"-Bathroom, "LR"-Living Room.}
    \label{I:vrhouse}
\end{figure}

Furthermore, we designed a virtual household environment (as seen in Fig.~\ref{I:vrhouse}), composed of 4 different sections: Kitchen (K), Living-Room (LR), Dining-Room (DR) and Bathroom (BR). In other words $\mathcal{L} = \left\{ \text{K}, \text{LR}, \text{DR}, \text{BR}\right\}$. Each section contains objects specific of that section (e.g. "Tooth-brush" is contained in the "Bathroom" area) as well as a number of common objects that can be found in multiple sections of the environment (e.g. "Cup" can be found in "Kitchen", "Living-Room", "Dining-Room").

\subsection{One-shot recognition (OSR) task}

In the one-shot recognition (OSR) task we evaluate the performance of OMCL in the recognition of actions when provided a single training demonstration of each class to create the associated motion concepts. We asked 10 participants to perform, on the virtual household environment, two demonstrations of (randomly-ordered) 22 action classes, after a tutorial period of adaptation to the VR setting. Each action was recorded for 6 seconds, storing the motion data, from the VR headset and motion controllers, and the contextual data (object and location information) of the performance. We provided to the participants no information regarding which objects to use or where to perform the action. The complete list of actions selected for this task is presented in Table~\ref{T:Motions}, along with the most common objects used in the performances and their most common locations in the household environment.

\begin{table}[]
\centering
\caption{List of action classes performed for the OSR task, along with the most common objects used in the performances and their most common locations in the household environment (following the nomenclature of Fig.~\ref{I:vrhouse}).}
\label{T:Motions}
\begin{tabular}{@{}lcc@{}}
\toprule
\multicolumn{1}{c}{Motion Class} & Location & Objects                    \\ \midrule
Bow                           & All      & None                       \\
Comb hair                        & BR       & Hairbrush                 \\
Cut                          & K        & Knife, Apple, Banana, Pear \\
Drink                        & All      & Mug, Glass, Bottle         \\
Eat at Table                  & DR       & Knife, Fork, Chopsticks    \\
Fry                           & K        & Frying Pan                 \\
High-Five                        & All      & Hand                       \\
Hug                          & All      & Body                       \\
Knock on door                 & LR       & Door                       \\
Pet                          & DR, LR   & Cat, Dog                   \\
Play Guitar                   & LR       & Guitar                     \\
Play Piano                    & DR       & Piano                      \\
Shake Hands                    & All      & Hand                       \\
Stir Pot                     & K        & Spoon, Pot                 \\
Sweep                         & K, LR    & Broom                       \\
Throw                         & All      & All                        \\
Vacuum clean                  & K, LR    & Vacuum-cleaner             \\
Wash Hands                    & BR       & Soap                       \\
Wash Plates                   & K        & Sponge, Dish               \\
Wash Window                   & K        & Sponge                     \\
Wave                           & All      & None                       \\
Wring Sponge                     & K     & Sponge                     \\ \bottomrule
\end{tabular}
\end{table}

The action classes were chosen due to their simplicity, as complex manipulation of objects in a virtual environment is difficult, and the fact that participants could performed them stationary, to minimize the discomfort of locomotion in virtual space. Moreover, we selected actions with very similar motion patterns but distinct object and location contexts (e.g. "Wash Hands"/"Wash Plates" and "Wave"/"Wash Window" actions) and actions with highly variant motion patterns, object and location contexts (e.g. "Throw" action).

\begin{figure*}
\begin{subfigure}{.32\textwidth}
  \centering
  \includegraphics[width=.99\linewidth]{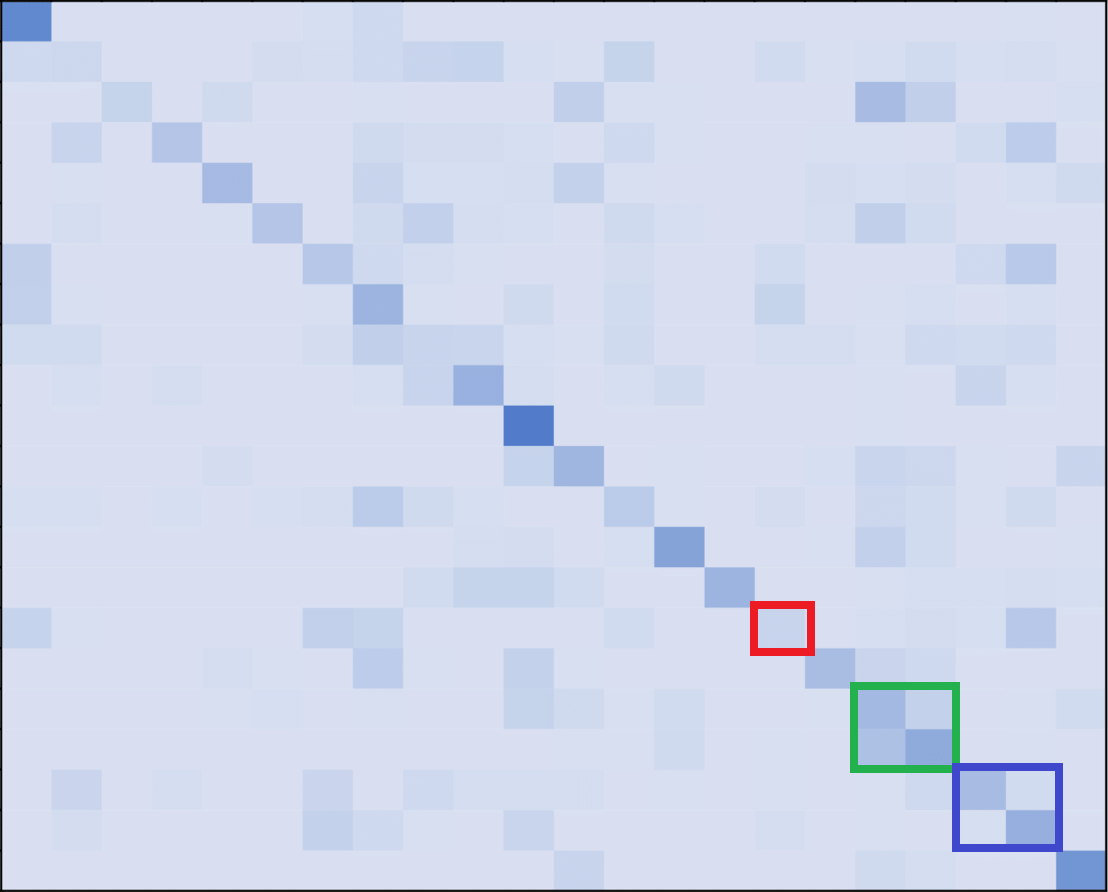}
  \caption{}
  \label{I:CM:HMM}
\end{subfigure} \quad
\begin{subfigure}{.32\textwidth}
  \centering
  \includegraphics[width=.99\linewidth]{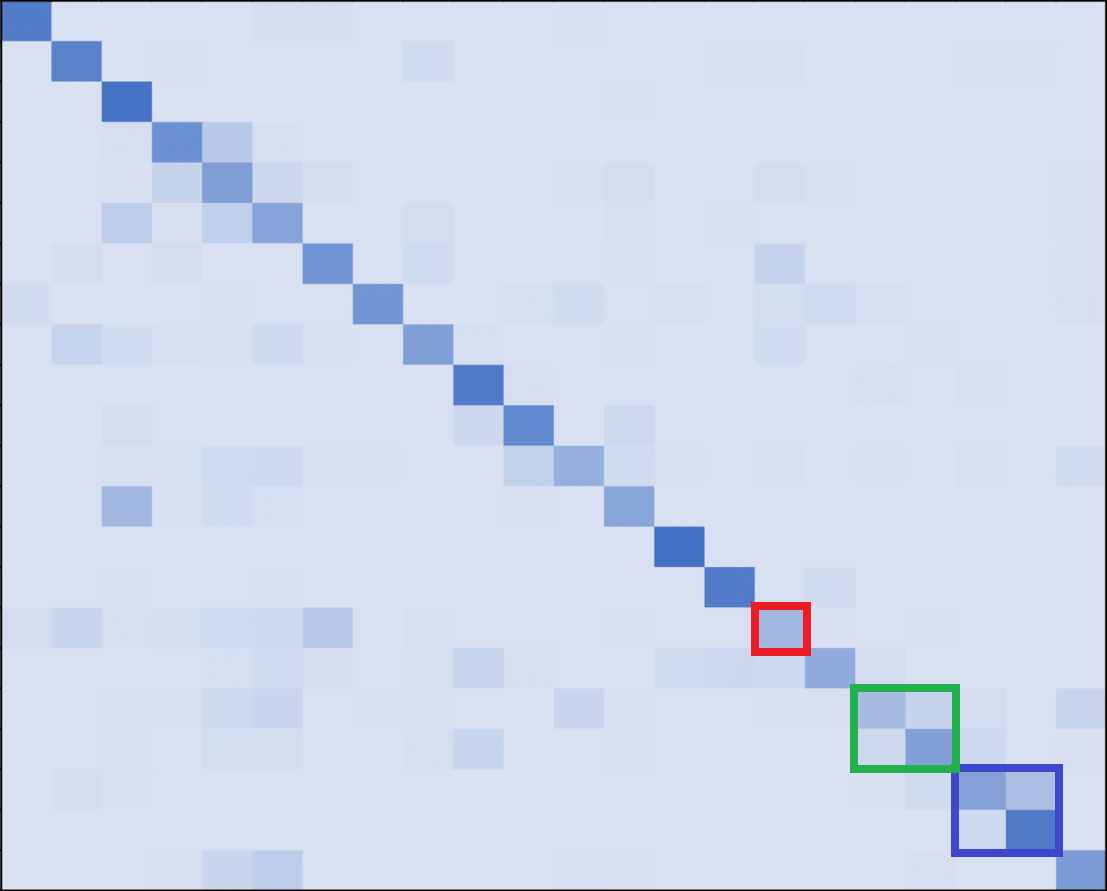}
  \caption{}
  \label{I:CM:OMCLNC}
\end{subfigure} \quad
\begin{subfigure}{.32\textwidth}
  \centering
  \includegraphics[width=.99\linewidth]{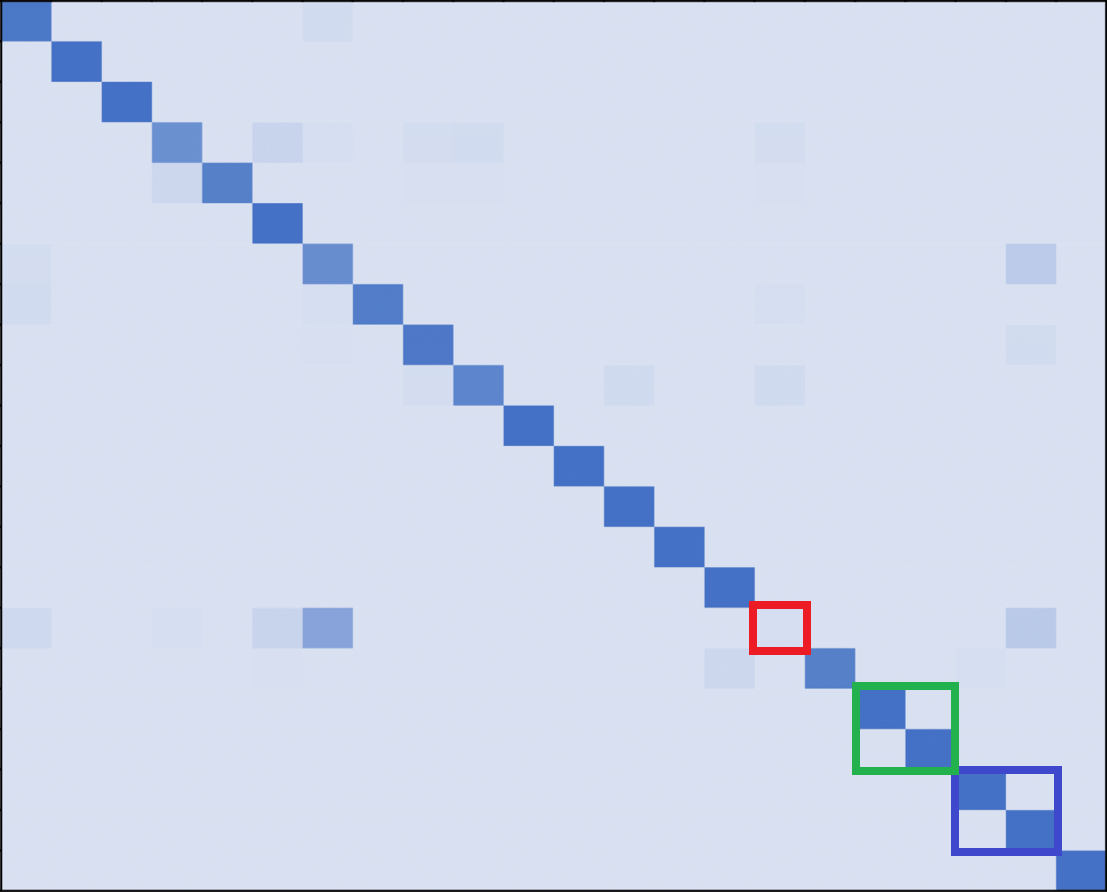}
  \caption{}
  \label{I:CM:OMCL}
\end{subfigure}%
 \caption{Confusion Matrices of the GMM-HMM (\ref{I:CM:HMM}), OMCL-N (\ref{I:CM:OMCLNC}) and OMCL (\ref{I:CM:OMCL}) algorithms on the one-shot recognition task. We highlight the accuracy on the "Throw" action class (red), the accuracy on the "Wash Hands" and "Wash Dishes" classes (green) and the accuracy on the "Wash Window" and "Wave" classes (blue).}
    \label{I:OSR_CM}
\end{figure*}

We optimize the values of the $(k_{\lambda,0},k_{\rho,0})$ parameters of OMCL by grid-search validation, training with one random sample of each action class and evaluating the remaining samples in the training partition of the dataset. The training procedure is repeated 10 times per tuple of parameter values and the optimized values are selected based on the total accuracy of the model. The $\Delta_{\mathcal{C}}$ parameter is optimized following the same grid-search procedure: fixing the values of $(k_{\lambda,0}, k_{\rho,0})$ obtained previously, we build a motion concept from a single randomly-selected training sample of each action class. Subsequently, we evaluate the number of times OMCL assesses the test samples (provided without explicit class labels) as examples of the correct, corresponding, motion concept. The final parameter values are presented in Table~\ref{T:params}.

\begin{table}[]
\centering
\caption{Optimized parameter values of the OMCL algorithm.}
\label{T:params}
\begin{tabular}{@{}lc@{}}
\toprule
Parameter                    & Value \\ \midrule
$k_{\lambda,0}$                   & 0.005 \\
$k_{\rho,0}$                   & 0.05  \\
$\Delta_{\mathcal{C}}$                   & 0.9   \\ \bottomrule
\end{tabular}
\end{table}

The performance of the OMCL algorithm is evaluated against a Gaussian Emission Hidden Markov Model (GMM-HMM), optimized through the same training procedure, yet resorting only to the motion data of the recorded actions. Using the total accuracy of the model as the selection criteria, the optimized number of hidden states in the model is $h_{\text{HMM}} = 16$ and the optimized number of components in the GMMs is $k_{\text{GMM}}= 3$. Moreover, to fairly compare both algorithms, we include in the evaluation procedure a modified OMCL model (OMCL-N), in which we neglect the contribution of the contextual features (object and location information) to the recognition cost (Eq.~\ref{EQ:total_comp_cost}). In other words, the motion concepts in OMCL-N are built solely considering motion data. In the OSR task, the recognition rates of the GMM-HMM algorithm, OMCL-N and OMCL algorithms in the test partition of the dataset are presented in Table~\ref{T:OSR}. Moreover, their confusion matrices are presented in Fig.~\ref{I:OSR_CM}.

\begin{table}
\centering
\caption{Accuracy on the one-shot recognition task for the GMM-HMM, OMCL-N and OMCL algorithms.}
\label{T:OSR}
\begin{tabular}{@{}lll@{}}
\toprule
\multicolumn{1}{c}{GMM-HMM ($\%$)} & \multicolumn{1}{c}{OMCL-N ($\%$)} & \multicolumn{1}{c}{OMCL ($\%$)} \\ \midrule
\multicolumn{1}{c}{$37.6 \pm 21.2$}    &    \multicolumn{1}{c}{$68.8 \pm 19.7$}       &  \multicolumn{1}{c}{$90.5 \pm 20.8$}    \\ \bottomrule
\end{tabular}
\end{table}

 In the OSR task, the OMCL-N algorithm significantly outperforms the GMM-HMM algorithm, with an accuracy of $68.8\pm 19.7\%$ against $37.6\pm 21.2\%$. This result validates the methodology of solving the recognition problem not through the direct comparison of low-level joint data, which is prone to noise and measurement errors, but through the comparison of previously learned motion primitives, generalized from the data. Yet, due to the diversity of motions patterns for the same action, contextual information still seems fundamental for the one-shot recognition, as the regular OMCL algorithm significant out-performs both methods (with $90.5\pm 20.8\%$ accuracy rate).
 
 The difference in performance between the algorithms can also be verified by the confusion matrices shown in Fig.~\ref{I:OSR_CM}. OMCL-N (Fig.~\ref{I:CM:OMCLNC}) presents a significantly more diagonal matrix compared to the matrix of GMM-HMM (Fig.~\ref{I:CM:HMM}). Yet, the recognition of actions with similar motion patterns is still difficult, as both algorithms are not able to successfully distinguish between the "Wash Hands"/"Wash Dishes" actions (marked in green in Fig.~\ref{I:OSR_CM}) as well as between the "Wave"/"Wash-Window" actions (in blue in Fig.~\ref{I:OSR_CM}). The OMCL algorithm (Fig.~\ref{I:CM:OMCL}) shows a significant improvement in the recognition of the actions classes, indicated by the near-diagonal confusion matrix. Moreover, OMCL is able to distinguish between actions with similar motion patterns ("Wash Window"/"Wash Dishes", "Wave"/"Wash-Window") by taking into account the contextual information of the action (object and location data). However, OMCL is still unable to recognize the action "Throw" (marked in red in Fig.~\ref{I:OSR_CM}) due to the similarity of its motion pattern to the class "High-Five" and the variance of objects used and locations where it can be performed. Indeed, the consideration of contextual information in the recognition of the "Throw" action seems to worsen the accuracy performance of the algorithm in comparisson with the solely-motion-based version of OMCL. Yet, for the remaining action classes, the contextual information of the actions seem to play a fundamental role in the improvement of the recognition performance of OMCL.

\section{Conclusion}

In this paper we present motion concepts, a novel multimodal representation for human actions in a household environment, based on the kinematics of the demonstration, the objects interacted with during the action and the location where it was performed. Moreover, we present OMCL, a new algorithm for the creation and the recognition of motion concepts from demonstrations provided by human users.

We evaluated OMCL on a one-shot recognition task, which showed that the motion concept representation proposed is suitable to be used in action recognition from a single demonstration. Moreover we attest to the importance of contextual information of an action to recognize actions with similar motion patterns. We plan to further evaluate the algorithm on an online motion concept learning task.

The question of learning rich representations of actions in an environment is an ever-evolving subject. We plan to develop further work on the representation of actions performed by multiple agents and the extension of the motion concept representation for task learning. We believe that action representations are fundamental tools for attaining a profound understanding of human behavior in an environment and, ultimately, for the widespread use of artificial agents in household environments.
